\documentclass[11pt]{article}

\usepackage[final]{acl}

\usepackage{times}
\usepackage{latexsym}
\usepackage{booktabs}

\usepackage[T1]{fontenc}

\usepackage[utf8]{inputenc}

\usepackage{microtype}

\usepackage{inconsolata}

\usepackage{graphicx}

\title{ArabicDialectHub: A Cross-Dialectal Arabic Learning Resource and Platform}

\author{Salem Lahlou \\
  Mohamed Bin Zayed University of Artificial Intelligence  \\
  UAE \\
  \texttt{salem.lahlou@mbzuai.ac.ae}}

\begin{document}
\maketitle
\begin{abstract}
We present ArabicDialectHub, a cross-dialectal Arabic learning resource comprising 552 phrases across six varieties (Moroccan Darija, Lebanese, Syrian, Emirati, Saudi, and MSA) and an interactive web platform. Phrases were generated using LLMs and validated by five native speakers, stratified by difficulty, and organized thematically. The open-source platform provides translation exploration, adaptive quizzing with algorithmic distractor generation, cloud-synchronized progress tracking, and cultural context. Both the dataset and complete platform source code are released under MIT license. Platform: \url{https://arabic-dialect-hub.netlify.app}.
\end{abstract}

\section{Introduction}

Arabic, spoken by over 400 million people across diverse regions, exhibits significant dialectal variation that creates substantial barriers to cross-dialectal communication. While Modern Standard Arabic (MSA) serves as a lingua franca in formal contexts, daily communication relies heavily on regional dialects that differ significantly in lexicon, phonology, and syntax. Moroccan Darija, in particular, stands at considerable linguistic distance from both MSA and other Arabic dialects, incorporating substantial Berber, French, and Spanish influences.

Despite the practical importance of cross-dialectal communication, learning resources remain scarce. Most Arabic language learning platforms focus exclusively on MSA \cite{rani2021computer}, leaving dialect speakers without adequate tools to learn other varieties. Existing dialectal resources \cite{bouamor2018madar,zaidan2011arabic,padic}, while valuable for computational linguistics research, are primarily designed for NLP applications rather than language learning. This gap is particularly acute for Darija speakers seeking to communicate with speakers of Levantine or Gulf varieties.

We present ArabicDialectHub, comprising two complementary contributions. First, we introduce a curated collection of 552 phrases across six Arabic varieties: Moroccan Darija, Lebanese, Syrian, Emirati, Saudi, and MSA. The collection was generated using large language models and validated by five native speakers (three Moroccan Arabic and two Lebanese Arabic speakers). Phrases are stratified by difficulty level (beginner, intermediate, advanced) and organized into thematic categories covering communication scenarios from daily greetings to complex idiomatic expressions. Second, we present an open-source interactive learning platform demonstrating the resource's practical utility through multiple learning modalities: a translation hub for phrase exploration, an adaptive quiz system with intelligent distractor generation, progress tracking with cloud synchronization, and cultural context cards highlighting regional sensitivities.

Our contributions address three critical gaps. Unlike large-scale dialectal corpora designed for NLP research, our collection is explicitly structured for pedagogical use with difficulty stratification and contextual usage notes. Unlike existing Arabic learning applications focusing exclusively on MSA, our platform centers on cross-dialectal learning with Darija as default source. Finally, unlike static datasets, we provide a fully functional interactive system validating the resource's utility for real-world language learning. Both the phrase collection and complete platform codebase are released under MIT license.

\section{Related Work}

\subsection{Arabic Dialect Corpora}

Significant efforts document Arabic dialectal variation. The MADAR project \cite{bouamor2018madar} provides parallel translations of 2,000 sentences across 25 Arab city dialects along with MSA, English, and French, accompanied by a lexicon covering 1,045 concepts. The Arabic Online Commentary (AOC) dataset \cite{zaidan2011arabic} pioneered collection of naturally-occurring dialectal Arabic, harvesting 52 million words from online news comments with 108,000 sentences manually annotated for dialect identification. The Parallel Arabic DIalectal Corpus \citep[PADIC;][]{padic} contributes 2,000 parallel sentences across six dialects including Moroccan, Algerian, Tunisian, Palestinian, and Syrian varieties. Other works include \citet{multidialectal}.

For Moroccan Darija specifically, recent developments include the Darija Open Dataset (DODa) \cite{o2024darija} with approximately 150,000 Darija-English entries, and the Atlaset dataset \cite{atlasia2024atlaset} containing over 155 million tokens for model pretraining. The DarijaBanking corpus \cite{s2024darijabanking} demonstrates LLM-assisted dataset creation viability, using GPT-4 for initial translation with subsequent validation by five native speakers. While these corpora provide valuable resources for computational linguistics, they primarily serve NLP tasks rather than language learning.

\subsection{Cross-Dialectal Learning Tools}

Despite abundant Arabic learning applications, few address cross-dialectal communication. Mainstream platforms (Duolingo, Rosetta Stone, Pimsleur) focus exclusively on MSA, leaving dialect learners underserved. Recent work on cross-dialectal Arabic translation \cite{bei2025crossdialectal} evaluates large language models on MADAR and QADI datasets, demonstrating growing technological capability but limited deployment in learner-facing applications. Computer-assisted language learning (CALL) research for Arabic \cite{h2024effectiveness,b2025integrating} predominantly addresses MSA acquisition, with AI tools struggling on dialects \cite{suherman2024leveraging}.

\subsection{LLM-Assisted Dataset Creation}

The use of large language models for linguistic dataset creation has gained acceptance in NLP research. Comprehensive surveys document hundreds of LLM-generated datasets across languages and domains \cite{liu2024datasets}. Critical to this approach is rigorous validation: the CLIcK benchmark for Korean \cite{click} employed four native speakers to validate all samples, while LAG-MMLU for Latvian and Giriama \cite{lag} used native speaker curation ensuring linguistic and cultural relevance despite persistent automatic translation errors. These precedents establish that LLM-assisted generation with native speaker validation represents viable methodology for creating linguistic resources, particularly for lower-resourced varieties.

\section{Methodology}

\subsection{Phrase Collection}

\subsubsection{Collection Strategy and Design}

The phrase collection was designed to serve practical cross-dialectal communication needs while maintaining pedagogical utility. Our selection criteria prioritized frequency (common everyday expressions), utility (practical value for real-world scenarios), and cultural relevance (appropriateness across diverse contexts). The collection spans 18 thematic categories. To accommodate learners at different proficiency levels, phrases were stratified into three difficulty tiers.  Additionally, 400 daily conversation sentences provide extended practice with natural dialogues. 

\subsubsection{LLM-Assisted Generation and Validation}

We useed large language models (Claude 3.5 and GPT-4) to generate initial translations, followed by rigorous human validation. Our prompt engineering approach specified detailed dialect characteristics for each target variety, provided contextual information about appropriate register and formality levels, and requested natural conversational translations rather than literal word-for-word renderings. The generation process proceeded iteratively, with initial translations undergoing consistency checking across dialects to identify outliers. Multiple generation attempts were compared for quality assessment, with validators selecting or synthesizing the most natural options.

The 552 phrases underwent review by five native Arabic speakers: three speakers of Moroccan Arabic (Darija) and two speakers of Lebanese Arabic, all fluent in multiple Arabic varieties and MSA. Validators reviewed phrases independently, focusing on naturalness (authentic native speech), accuracy (semantic equivalence), and cultural appropriateness (suitability for indicated contexts).

\subsection{Platform Development}

To demonstrate the practical utility of our phrase collection, we developed ArabicDialectHub, an open-source web application providing multiple interaction modalities for cross-dialectal Arabic learning. The platform serves as both proof-of-concept for the resource's pedagogical value and a contribution to the dialectal Arabic learning ecosystem.

\subsubsection{System Architecture}

The platform employs a modern web architecture optimized for responsive cross-platform access. The frontend utilizes React 18 and TypeScript for type-safe component development with efficient client-side rendering. 
The backend leverages two specialized cloud services. Clerk handles authentication, supporting email/password credentials with secure session management. This separation of authentication concerns from data management provides security best practices and simplifies user account management. Supabase provides the data layer through a PostgreSQL database with built-in real-time synchronization capabilities and row-level security policies ensuring users can only access their own progress data while maintaining shared read access to the phrase collection.

The database schema comprises three primary tables. The \texttt{phrases} table stores all 552 phrases with complete metadata, synchronized from source JSON files during deployment. The \texttt{phrase\_progress} table tracks individual user mastery status for each phrase, recording correctness counts, mastery flags, and last review timestamps. The \texttt{quiz\_attempts} table logs all quiz sessions with scores, question counts, configuration parameters (source/target dialects, difficulty filters), and completion timestamps, enabling progress analysis and learning analytics. The platform is deployed on Netlify CDN with continuous integration from GitHub, ensuring automatic updates as the codebase evolves.

\subsubsection{Core Learning Features}

\textbf{Translation Hub} serves as the primary exploration interface for discovering and learning phrases across Arabic dialects. The hub displays three randomly-selected unmastered phrases simultaneously, encouraging focused attention while maintaining variety. This limited display prevents overwhelming learners while ensuring adequate exposure to new content. Users can expand phrase cards through accordion components to view translations across all six dialects, facilitating cross-dialectal comparison and pattern recognition. Each phrase includes the Darija original in Arabic script with Latin transliteration, literal English translation, and complete translations for Lebanese, Syrian, Emirati, Saudi, and MSA varieties. Each dialectal translation provides Arabic script, romanization for pronunciation guidance, and usage notes explaining contextual appropriateness. The mastery tracking system employs a simple one-click approach where users mark phrases as "mastered" when confident in their understanding. Once marked, phrases are removed from the default rotation but remain accessible through a ``show mastered'' toggle, allowing review without cluttering the learning interface. All mastery status changes synchronize immediately to Supabase, enabling seamless cross-device learning. The hub includes search functionality for targeted phrase lookup by text content, and filtering options by category and difficulty level. A progress indicator displays real-time mastery percentage, providing immediate feedback on learning advancement.

\textbf{Quiz System} provides active recall practice essential for language retention through two question types with varying cognitive demands. Multiple-choice questions present a phrase in the source dialect with four answer options in the target dialect. The distractor generation algorithm identifies phrases with similar lexical or phonological characteristics from maintained distractor banks.  Word-ordering questions present a phrase in the source dialect and challenge users to arrange shuffled words into the correct target dialect sequence. This question type tests syntactic understanding and productive competence rather than mere recognition, requiring learners to actively construct grammatically correct sentences. The shuffling algorithm ensures that word order is sufficiently scrambled to prevent pattern-based guessing. The system provides immediate feedback after each question with green highlights for correct answers and red highlights for incorrect responses, displaying the correct answer. 

\textbf{Progress Tracker} visualizes learning metrics across multiple dimensions to support metacognitive awareness and motivation. The dashboard displays overall statistics including total phrases mastered (both absolute count and percentage of the 552-phrase collection) and average quiz performance across all attempts. 

\textbf{Cultural Context Cards} acknowledge that effective cross-dialectal communication requires cultural competence alongside linguistic knowledge. Rather than embedding cultural information solely within individual phrases, dedicated thematic cards provide broader context about regional differences in social norms, communication styles, and cultural sensitivities. Each cultural card presents key points with concrete examples, regional differences across the five dialects, and practical tips for real-world usage. For instance, the greetings card explains that while formal greetings are universally important, the expected response patterns vary (Syrian Arabic often includes more elaborate well-wishing phrases, while Gulf varieties may incorporate more formal religious expressions). This contextualization prevents social mistakes and enhances pragmatic competence.

\section{Discussion}

\subsection{Resource Contribution}

The ArabicDialectHub phrase collection addresses a critical gap in cross-dialectal Arabic resources. While existing corpora such as MADAR and PADIC provide valuable parallel data for computational research, our collection offers explicit pedagogical structuring through difficulty stratification, enabling progressive learning, unlike research corpora treating all data uniformly. Second, rich contextual metadata including usage notes, formality indicators, and cultural sensitivities directly supports learner needs beyond mere translation equivalence. 

\subsection{Platform Validation}

The platform demonstrates that our phrase collection effectively supports interactive language learning. The translation hub validates that the resource structure facilitates browsing and discovery. The quiz system's successful distractor generation confirms sufficient lexical diversity for automated assessment without external word lists. Progress tracking proves feasible through the structured data model. The open-source release establishes a blueprint for dialectal learning applications beyond Arabic.

\section{Conclusion}

We have presented ArabicDialectHub, a comprehensive resource for cross-dialectal Arabic learning comprising a curated phrase collection and an interactive learning platform. The collection of 552 phrases across six Arabic varieties addresses the critical gap in pedagogically-oriented dialectal resources, particularly for Moroccan Darija speakers. Our methodology combining LLM-assisted generation with native speaker validation demonstrates a viable approach for efficient resource creation in lower-resourced language varieties. The open-source platform provides multiple learning modalities (browsing, quizzing, progress tracking, and cultural context) validating the resource's practical utility.

While acknowledging significant limitations in validation scope, empirical evaluation, and scale, we position this work as an enabling contribution. The complete dataset and platform code are publicly available under the MIT license at \url{https://github.com/saleml/arabic-dialect-hub}, with documentation supporting extension and adaptation. We welcome contributions from the research community and language learning practitioners to refine translations, expand dialect coverage, integrate audio resources, and conduct rigorous pedagogical evaluation. By lowering barriers to cross-dialectal Arabic learning and establishing open infrastructure for collaborative development, we aim to support both learner communities and future research on dialectal language education.

\section*{Limitations}

\textbf{Validation Coverage.} 
Native speaker validation was limited to Moroccan Darija (three validators) and Lebanese Arabic (two validators). Syrian, Emirati, and Saudi translations lack native speaker verification. Inter-annotator agreement metrics were not computed.
\\
\textbf{Dataset Scale and Scope.} 
At 552 phrases, the collection remains modest compared to research corpora like MADAR (2,000 sentences). Several domains including technical, medical, and professional vocabulary are absent, constraining utility for specialized communication needs.
\\
\textbf{Methodology Documentation.} 
Difficulty levels were assessed by Claude 3.5 without formal operationalization criteria. The choice of general-purpose LLMs over Arabic-focused models (e.g., Jais) was driven by availability rather than systematic comparison.
\\
\textbf{Evaluation.} 
No user studies or learning outcome assessments were conducted. Platform effectiveness for language acquisition remains unvalidated.
\\
\textbf{Modality.} 
The absence of audio recordings limits pronunciation learning and productive skill development.

\section*{Ethics Statement}

\textbf{LLM Use.} Claude 3.5 and GPT-4 generated initial translations, with all content undergoing mandatory native speaker validation. LLMs served as productivity tools rather than authoritative sources. Potential biases include systematic preference for formal registers in MSA-trained models.
\\
\textbf{Data Privacy.} Platform collects minimal user data: authentication credentials (Clerk-managed), learning progress (Supabase with row-level security), quiz histories. No PII beyond email retained.

\bibliography{custom}

\end{document}